\documentclass{svproc}
\usepackage{makeidx}  

\usepackage{mathtools}
\usepackage{graphicx}
\graphicspath{ {./images/} }
\usepackage{xcolor}
\definecolor{darkGray}{RGB}{135,135,134}
\definecolor{CodeGreen}{rgb}{0,0.6,0}
\definecolor{CodeMauve}{rgb}{0.58,0,0.82}
\usepackage{algorithm}
\usepackage{algpseudocode}
\usepackage{listings}
\usepackage[hidelinks]{hyperref}

\begin{document}
\frontmatter          
\pagestyle{headings}  
\addtocmark{Human Emergency Detection during Autonomous Hospital Transports} 

\mainmatter              
\title{Human Emergency Detection during Autonomous Hospital Transports}

\titlerunning{Human Emergency Detection}  
\author{Andreas Zachariae\inst{1} \and Julia Widera\inst{1} \and Frederik Plahl\inst{1} \and Björn Hein\inst{1} \and Christian Wurll\inst{1}}
\authorrunning{Andreas Zachariae et al.} 
\tocauthor{Andreas Zachariae, Julia Widera, Björn Hein, Christian Wurll}

\institute{Karlsruhe University of Applied Sciences, Germany\\
\email{Andreas.Zachariae@h-ka.de},\\ Homepage:
\url{www.h-ka.de/iaf/iras}}
\maketitle

\begin{abstract}
Human transports in hospitals are labor-intensive and primarily performed in beds to save time. This transfer method does not promote the mobility or autonomy of the patient. To relieve the caregivers from this time-consuming task, a mobile robot is developed to autonomously transport humans around the hospital. It provides different transfer modes including walking and sitting in a wheelchair. The problem that this paper focuses on is to detect emergencies and ensure the well-being of the patient during the transport. For this purpose, the patient is tracked and monitored with a camera system. OpenPose is used for Human Pose Estimation and a trained classifier for emergency detection. We collected and published a dataset of 18,000 images in lab and hospital environments. It differs from related work because we have a moving robot with different transfer modes in a highly dynamic environment with multiple people in the scene using only RGB-D data. To improve the critical recall metric, we apply threshold moving and a time delay. We compare different models with an AutoML approach. This paper shows that emergencies while walking are best detected by a SVM with a recall of \mbox{95.8 \%} on single frames. In the case of sitting transport, the best model achieves a recall of \mbox{62.2 \%}. The contribution is to establish a baseline on this new dataset and to provide a proof of concept for the human emergency detection in this use case.
\keywords{Autonomous Hospital Transport, Fall Detection, Human Pose Estimation, Dataset}
\end{abstract}

\section{Introduction}
\subsection{Autonomous Hospital Transports}
\begin{figure}[h]
\includegraphics[width=\textwidth]{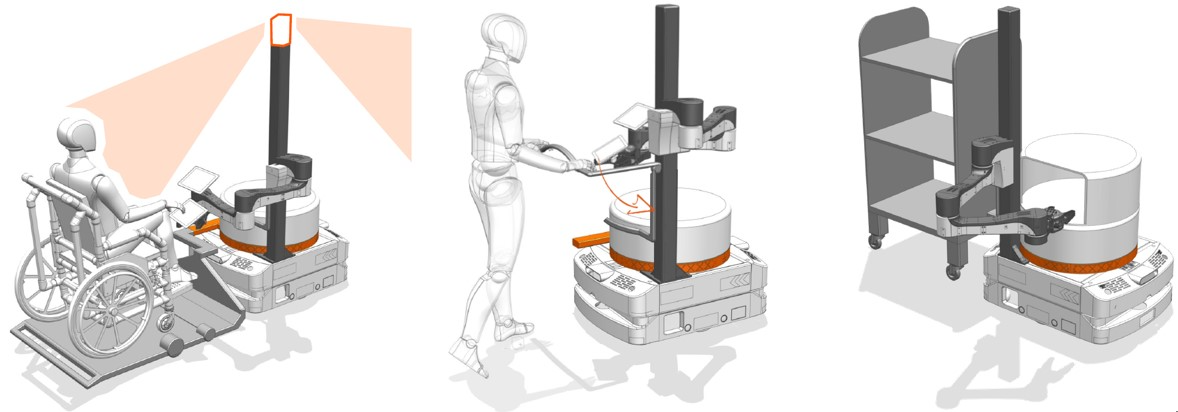}
\caption[]{Multi-mobility methods of PeTRA: Wheelchair transport (left), sensory coupling (center) and material transport (right).}
\centering
\label{fig:multi_mobility_methods}
\end{figure}

Transporting patients from ward to examination is part of the daily routine in a hospital. This manual transport is currently also performed by trained nurses, who are absent from the ward during this time and cannot perform nursing activities. To save time, patients are moved through the hospital mainly in beds, even though some of them are able to walk by themselves. The currently used transfer method neither promotes the patients' mobility nor allows them to decide for themselves whether and how they want to walk. The BMBF-funded research project of which this work is a part, is developing a Person Transfer Robot Assistant (PeTRA) to solve these problems. The goal is to relieve caregivers from time-consuming and labor-intensive person transfers in order to have more time for qualitative care \cite{petra-konsortium_personen-transfer_2022}. For this task, an autonomous mobile robot was developed in close collaboration with three hospitals. PeTRA offers different transfer modes, which each patient can use individually as needed, see figure  \ref{fig:multi_mobility_methods}. Autonomous patient transport is realized with sensory coupling for free walking or with the support of a rollator (center) and a modular platform for wheelchairs (left). This platform is used for safe wheelchair transport and can be coupled. In addition, an integrated robotic arm performs service tasks (right). These include, for example, the transport of drugs or blood samples between the storage area and the wards..

\subsection{Human Emergency Detection}
To ensure the patient's well-being during transport, a depth camera monitors the patient's condition. Many unforeseen situations can occur during a transport, not all of which are dangerous and require an emergency call. Some activities are part of everyday actions but are very difficult to distinguish from critical situations, such as kneeling down to tie a shoe or bending over to pick up an object. However, in the event of a fall or unconsciousness, help should be requested immediately. In addition to normal walking, transport modes with rollators and wheelchairs require reliable detection even when people are seated or partially obscured. Especially difficult is the detection of unconscious patients in wheelchairs, even for human caregivers. An additional requirement of the use case is to prevent the robot from being classified as a medical device that could monitor the patient's heart rate or blood pressure. This excludes the use of portable pulse oximeters or similar devices. Accelerometers are also not used in this paper because each patient would need to be outfitted prior to transport, but the goal is to minimize transfer time and caregiver tasks. The emergency detection in this paper uses only RGB images and corresponding depth images.

Previous research has primarily considered static scenes for fall detection in nursing homes. A systematic review was conducted by \cite{ren_research_2019}. One approach is to use inertial data from accelerometers, either as an additional device \cite{yacchirema_fall_2019} or from wearables such as a smartphone \cite{shahzad_falldroid_2019} or smartwatch \cite{kostopoulos_increased_2015}. Another common use case is static detection from ceiling-mounted cameras \cite{de_miguel_home_2017}\cite{fan_fall_2017} or acoustic \cite{droghini_human_2018} and radar sensors \cite{tian_rf-based_2018}. For camera-based detection, human pose estimation methods such as OpenPose \cite{chen_fall_2020}\cite{lin_framework_2021}\cite{wang_fall_2020}\cite{huang_video-based_2018} or AlphaPose \cite{ramirez_fall_2021} are used for feature generation. Mobile robots are also used to detect fallen persons with cameras. These robots are applied as patrol-robots and detect already fallen persons lying on the ground \cite{maldonado-bascon_fallen_2019}\cite{sumiya_mobile_2015}\cite{volkhardt_fallen_2013}.

\begin{table}[h]
\caption{The relevant classes of patient transports in hospitals}
\label{table:classes}
\centering
\begin{tabular}{llllll}
\hline\noalign{\smallskip}
\multicolumn{1}{c}{\textbf{Class Name}} & \multicolumn{1}{c}{ID} & \multicolumn{1}{c}{Trigger actions} & \multicolumn{1}{c}{Reaction} \\
\noalign{\smallskip}
\hline
\noalign{\smallskip}
\textbf{Normal} & 0 & \begin{tabular}[c]{@{}l@{}}Walking, sitting in wheelchair, \\ everyday situations, pushing a rollator\end{tabular} & Transport as planned \\
\noalign{\smallskip}
\textbf{Emergency} & 1 & \begin{tabular}[c]{@{}l@{}}Fall, stumbling with rollator, \\ unconscious in wheelchair\end{tabular} & \begin{tabular}[c]{@{}l@{}}Transport stopped and \\ emergency call to nurse\end{tabular} \\
\noalign{\smallskip}
\textbf{Pause} & 2 & \begin{tabular}[c]{@{}l@{}}Patient too far away, standing up \\ from wheelchair, crushing hazard\end{tabular} & Transport paused \\
\hline
\end{tabular}
\end{table}

In contrast, this work considers different transportation modes of a mobile robot in a highly dynamic environment with multiple people in one scene using only RGB-D data. Difficulties such as occlusion by rollators or wheelchairs and scenarios like unconsciousness or crushing hazards pose additional challenges. This makes the use case increasingly complex. For real patient transport in hospitals, three classes are relevant: \textit{Normal} transport as the default case, \textit{Emergency} mode when a critical situation occurs, and the \textit{Pause} case when there is a problem but transport can continue, see table \ref{table:classes}.

\section{Dataset}

The PeTRA dataset consists of over 18,000 single images from 200 videos as a single label, multiclass classification problem. It is a rich training source for the use case of a moving robot in a highly dynamic environment with multiple people in a scene. There are three different use cases for this dataset, corresponding to the transportation modes of the PeTRA robot. The distribution of the images in each class is shown in table \ref{table:dataset}. The first transport mode is only with self-walking patients, this includes patients with rollator who can be partially occluded. This set consists only of the binary output classes \textit{Normal} and \textit{Emergency}. The second mode is for transportation with wheelchair only and the third mode is a combination of both datasets for transports where both scenarios occur.

\begin{figure}[H]
\centering
\includegraphics[width=0.92\textwidth]{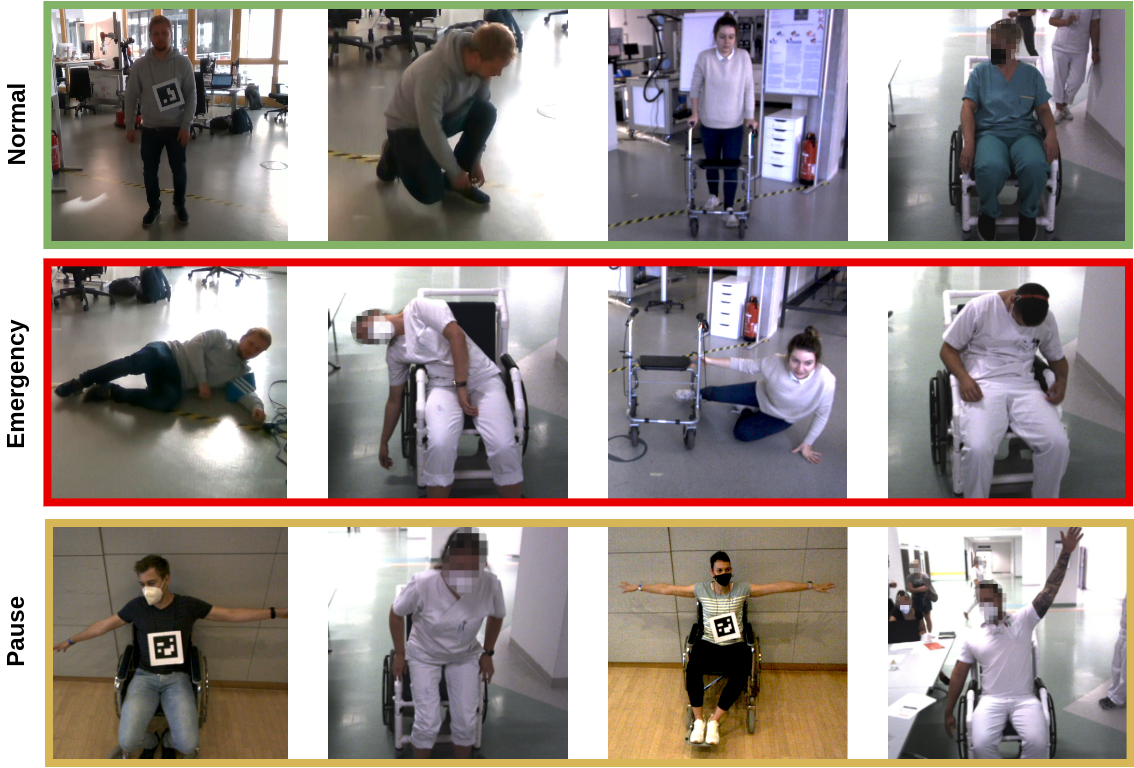}
\caption[]{Example images from the \textit{PeTRA Emergency Detection Dataset}}
\label{fig:dataset_example_images}
\end{figure}

\begin{table}[h]
\caption{The PeTRA dataset with its three application modes}
\label{table:dataset}
\centering
\begin{tabular}{llllll}
\hline\noalign{\smallskip}
\multicolumn{1}{c}{\textbf{Class Name}} & \multicolumn{1}{c}{ID} & \multicolumn{1}{c}{Walking} & \multicolumn{1}{c}{Wheelchair} & \multicolumn{1}{c}{Combined} & \multicolumn{1}{c}{\%} \\
\noalign{\smallskip}
\hline
\noalign{\smallskip}
\textbf{Normal} & 0 & 7,864 & 4,704 & 12,568 & 68 \% \\
\textbf{Emergency} & 1 & 2,550 & 980 & 3,530 & 19 \% \\
\textbf{Pause} & 2 & 0 & 2,304 & 2,304 & 13 \% \\
\noalign{\smallskip}
\hline
\noalign{\smallskip}
\textbf{$\Sigma$} &  & 10,414 & 7,988 & 18,402 & 100 \% \\
\hline
\end{tabular}
\end{table}
All of the scenes are partially static and dynamic, in which the robot moves in front of the person. The use case requires a multi-person detection with tracking over frames, therefore the number of people simultaneously visible in the scene ranges from 1 to 5. There are 20 unique people with an equal distribution of men and women. The locations differ from scenes in the laboratory with and without daylight as well as scenes from hospital floors with only artificial lighting, see figure \ref{fig:dataset_example_images} for examples. It was captured with two different stereo cameras. The first is a \textit{Roboception rc\textunderscore visard 160 color} which has a depth range from 0.5 m to infinity. The second camera is an \textit{Intel RealSense D415} with a depth range from 0.5 m to 3 m.

Detailed labels with 22 classes were created manually. These include different stages of falling and specific everyday situations. The final output classes relevant for emergency detection are derived from these labels. For testing 5618 \mbox{(31 \%)} images are used. The other part is used for training with a 5-fold cross validation. The dataset split is made on video level, so that no image from the same video is in both sets. The test set also contains videos of people who do not appear in the training set. As a contribution of this paper, a subset of the PeTRA dataset with all rights for publication is released. A reduced version without the original images, but with all the keypoints from OpenPose and corresponding depth data, already filtered by the relevant patient, can be found on GitHub (\url{www.github.com/AndreasZachariae/PeTRA_Dataset_Human_Emergency_Detection}).

\section{Implementation}

\begin{figure}[h]
\includegraphics[width=\textwidth]{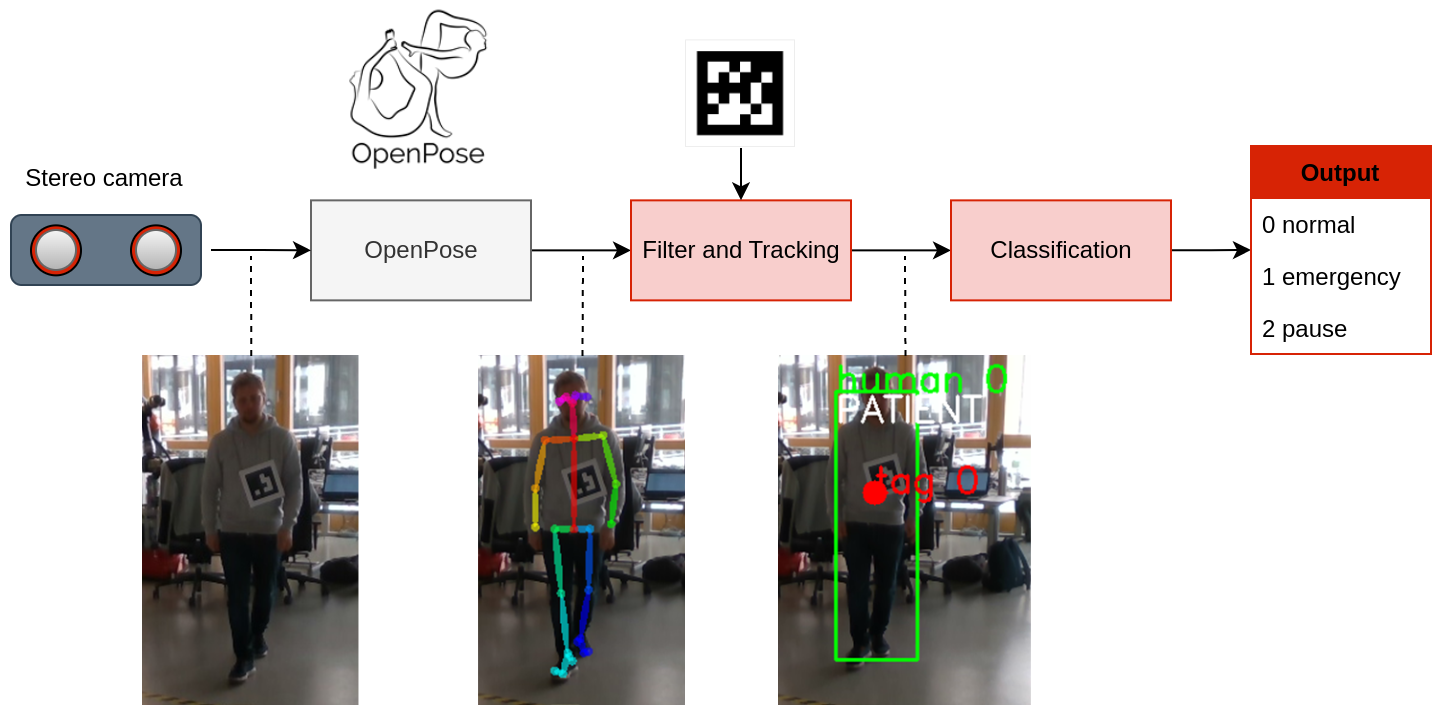}
\caption[]{Data processing pipeline from stereo camera to class output}
\centering
\label{fig:pipeline}
\end{figure}

Figure \ref{fig:pipeline} shows the data processing pipeline from stereo camera to class output. The whole pipeline is implemented using ROS2 Foxy \cite{macenski_robot_2022}. First the depth camera captures images while tracking the patient with a 2-axis pan-tilt unit. These images are processed by OpenPose for Human Pose Estimation. OpenPose is a CNN for real-time recognition of multiple human poses in a 2D image \cite{cao_openpose_2021}. OpenPose downscales all RGB images to 480x360 at 10 Hz and is configured to maximize detections. This highly increases both false and true positives and maximizes average recall. High recall for human detection is important because during a fall, the certainty of identifying a human decreases, but the patient should still be detected and tracked. The additional false positives are countered by downstream filtering. The extracted keypoints, which indicate the positions of joints and body parts of all detected humans, are filtered for the relevant patient who has to be transported. This is done by tracking the relevant patient over each frame and discarding all other keypoints. To increase accuracy, the patient can wear a visual marker such as ArUco \cite{garrido-jurado_automatic_2014}. It is also used to initially mark the relevant patient, as opposed to other people who may be in the field of view when the transport begins. In a control loop with the pixel position information of the detected person or the ArUco marker, the 2-axis pan-tilt unit moves the camera to point at the relevant patient and ensures he is centered in the image. The 2D keypoints of the patient are then matched with the depth image to get the corresponding distances of the body parts to the camera. This data is used to train the classifiers to determine one of the three output classes.

\section{Method}
\subsection{Training of recall-optimized classifier}

In Human Emergency Detection, an undetected emergency is critical and more significant than other misclassifications. The real application domain has highly unbalanced class distributions and the training data reflects this. The \textit{Normal} class is the default case and occurs most of the time, hence the accuracy of the classification is not a good quality score. For only self-walking patients, it is a single label binary classification problem where the amount of false negatives should be minimal, therefore the recall (\ref{eqn:recall}) should be maximized.
\begin{equation}
\label{eqn:recall}
\mbox{recall} = \frac{\mbox{TP}}{(\mbox{TP} + \mbox{FN})}
\end{equation}
In the case of wheelchair and combined transports, it is a single label multiclass classification problem where only the FNs of the \textit{Emergency} class are critical. The metric to optimize is defined as micro-averaged recall (\ref{eqn:recall-micro}).
\begin{equation}
\label{eqn:recall-micro}
\mbox{recall}_{\mbox{micro}} = \frac{\sum_{\mbox{classes}} \mbox{TP of class}}{\sum_{\mbox{classes}} \mbox{TP of class } + \mbox{ FN of class}}
\end{equation}
Three different model types are evaluated and compared with an AutoML approach. The single models are Random Forest (RF), Multilayer Perceptron (MLP), and Support Vector Machine (SVM), each with four hyperparameters that are optimized in a grid search with 5-fold cross validation.

\subsection{Decision threshold optimization}
\begin{figure}[h]
\includegraphics[width=\textwidth]{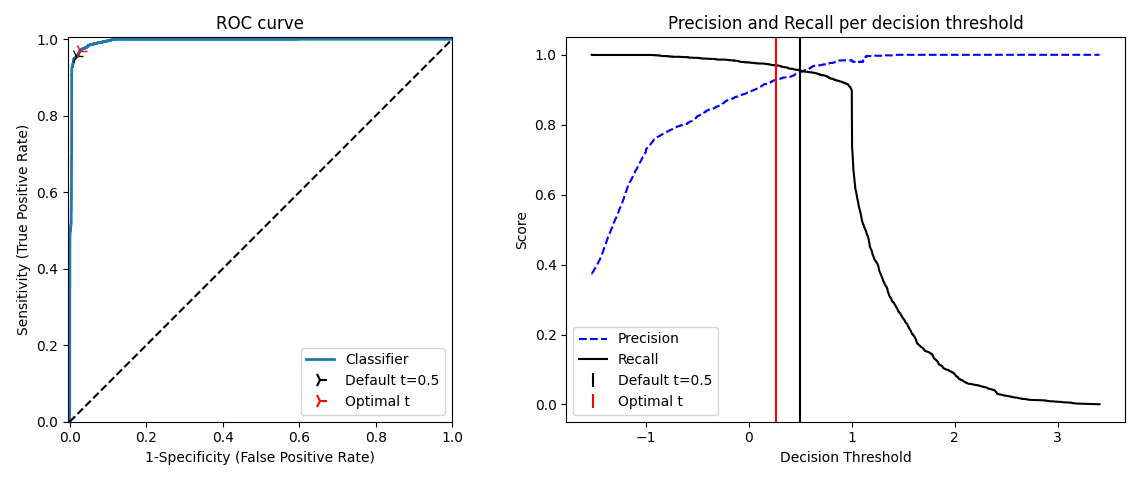}
\caption[]{Effect of threshold moving on ROC curve (left), Precision and Recall (right)}
\centering
\label{fig:threshold_moving}
\end{figure}

The selected models are capable of predicting a class probability instead of directly giving the class label. The default threshold in the binary case is 0.5. For the unbalanced classification problem of PeTRA, the default threshold can lead to lower performance, as described by \cite{brownlee_gentle_2020}. To take account for this, the decision thresholds of all models are optimized using the training data. In the self-walking binary case, the optimality criterion is the maximum of Youden's J statistic (\ref{eqn:J}) \cite{youden_index_1950}.
\begin{equation}
\label{eqn:J}
J = \frac{\mbox{TP}}{\mbox{TP} + \mbox{FN}} + \frac{\mbox{TN}}{\mbox{TN} + \mbox{FP}} - 1 = \mbox{sensitivity} + \mbox{specificity} - 1
\end{equation}
The ROC curve visualizes the relation of \(\mbox{sensitivity}\) over \(1 - \mbox{specificity}\). The optimal \(J\) value is the point closest to the upper left corner where  \(\mbox{sensitivity} = \mbox{specificity} =  1\). The optimal \(J\) value gives the optimal number of false negatives and false positives weighted by the geometric mean. In the precision-recall curve in figure \ref{fig:threshold_moving}, the optimal threshold is highlighted with a red line and the default threshold with a black line. This shows, that moving the threshold serves the goal of increasing the models recall value. 

\lstset{language=Python, caption={Method for threshold optimization}, label=lst:python_code}
\begin{lstlisting}[language=Python]
def get_optimal_threshold(self, y_scores, y_train):
    thresholds = numpy.arange(0, 0.5, 0.001)
    y_scores = [self.softmax(y) for y in y_scores]
    f1_scores = []
    for t in thresholds:
        y_pred = [numpy.argmax(y) if y[1] < t else 1 for y in y_scores]
        f1_scores.append(f1_score(y_train, y_pred, average="micro", labels=[1]))
    return thresholds[numpy.argmax(f1_scores)]
\end{lstlisting}
\smallskip
\smallskip
\smallskip

In multiclass classification, there is no single threshold to divide the class probabilities. The decision probabilities are one-hot encoded and, depending on the model, are not bound to 0-1. A softmax function is applied to map them to the range from 0 to 1. On these values, the class of the maximum value is selected, except if the value of the \textit{Emergency} class is greater than the defined threshold, then it is classified as an emergency. See algorithm \ref{lst:python_code} for details. This leads to fewer undetected emergencies, as this class is predicted although some other class would have a higher score. As a result, the recall on the \textit{Emergency} class can be improved. The optimal threshold value is determined by the maximum micro-averaged F1-score (\ref{eqn:f1-micro}).

\begin{equation}
\label{eqn:f1-micro}
F1_{\mbox{micro}} = 2 \times \frac{\mbox{recall}_{\mbox{micro}} \times \mbox{precision}_{\mbox{micro}}}{\mbox{recall}_{\mbox{micro}} + \mbox{precision}_{\mbox{micro}}}
\end{equation}

\subsection{Delay Time}

Analysis of the behavior of the best models during one complete fall results in the introduction of an additional delay hyperparameter. On the walking dataset, this best model is the SVM with a recall of \mbox{95.8 \%}. There are two problems in application: First, the classification is unstable during a fall. There is some fluctuation between the \textit{Normal} and \textit{Emergency} classes before it becomes reliably stable. The SVM model correctly detects all 21 falls from the test videos and takes a maximum of 442 ms to become stable (mean: \mbox{94 ms}, std: \mbox{160 ms}). This delay is still sufficient to call help in case of an emergency. The second problem comes from optimizing the classifier to maximize recall at the expense of precision: Every false positive results in a false emergency alarm. These FPs occur as outliers during normal transport. Both problems can be reduced by introducing a delay between the classification and the triggering of an emergency behavior. 

\begin{figure}[h]
\includegraphics[width=\textwidth]{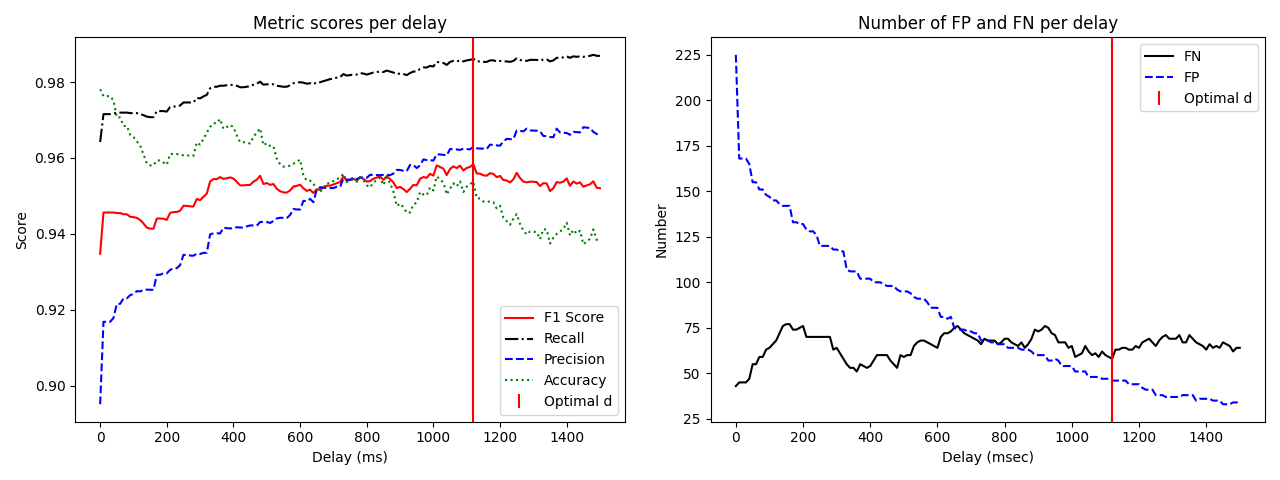}
\caption[]{Effect of the delay on metrics (left), FPs and FNs (right)}
\centering
\label{fig:optimal_delay}
\end{figure}

The optimal delay value is determined by the maximum micro-averaged F1-score (\ref{eqn:f1-micro}). The delay value is deduced from the training data as the time between the first true \textit{Emergency} frame until the first frame after which the prediction is stable for the rest of the video. The metrics are computed for each delay in the range from 0 to 1.5 s with 10 ms steps and the one with the best F1-score is selected as the optimal delay, see figure \ref{fig:optimal_delay}. This leads to lower FN and FP but not necessarily to a lower recall because the frames from the start of the delay to its end are counted towards the previously detected class, which is mostly the \textit{Normal} class, thus the proportions change.

\section{Results and Discussion}
\subsection{Comparison with AutoML approach}

This work uses hyperparameter tuning on a manually selected parameter range. To verify that this selection and parameter tuning is reasonable, the results are compared with Automated Machine Learning (AutoML) approaches. AutoML is an automated method for model algorithm selection, hyperparameter optimization, and model evaluation. The framework used for comparison is Auto-Sklearn 2.0 \cite{feurer_auto-sklearn_2022}. As goal metric the same micro-averaged recall (\ref{eqn:recall-micro}) is applied. The AutoML models are optimized for different time periods from 30 s to 24 h for each application mode. On average, the recall of all AutoML models on the test data is \mbox{5.3 \%-points} lower than the manually optimized models. The best AutoML model has \mbox{91.8 \%} recall on the walking test data which is \mbox{4.0 \%-points} lower than the best manual SVM model.

\subsection{Evaluation results}

The table \ref{table:evaluation_results} shows the evaluation results of the best models per model type for each application mode. The best model type for both, the binary and multiclass cases is for all application modes the SVM.

The threshold moving method could improve the recall for almost all model types except for the SVM on walking and RF on wheelchair data. Models with threshold moving are indicated with the suffix \textit{\_thresh} and the optimal threshold \(\hat{t}\) is given. On average, the recall of all models can be improved by \mbox{1.6 \%-points}. On the best model, which is the SVM on the walking dataset, the recall is already high at \mbox{95.8 \%} and is decreased slightly by \mbox{0.7 \%-points}. For the AutoML models, no threshold moving was applied, the value in the model name indicates the optimization time in seconds (\(25200 s = 7 h\)). 

With the introduction of the delay, the number of FNs and FPs is reduced. The optimal delay \(\hat{d}\) is always greater than zero. A recall of 10 ms is the smallest step and results in delaying the classification of an emergency for a single frame. This still leads to a significant reduction of the total number of FPs, expressed in the fraction \(\mbox{FP}_{\hat{d}} / \mbox{FP}\). On average, the FPs are reduced to \mbox{66.9 \%}, FNs are increased to \mbox{101.1 \%}, and the F1-score is improved by \mbox{0.3 \%-points}. In the best model, FPs are reduced to \mbox{21.4 \%}, FNs to \mbox{92.0 \%} and the F1-score is increased by \mbox{8.6 \%-points}. The effect on FNs is sometimes greater than 1 because the models were already optimized for recall. The delay is applied after the classification and does not effect the recall and F1 scores in the table.


\begin{table}[h]
\caption{Evaluation results for the best models per model type for each application}
\label{table:evaluation_results}
\centering
\begin{tabular}{llllllll}
\hline\noalign{\smallskip}
Application & Model name & \multicolumn{1}{c}{Recall} & \multicolumn{1}{c}{F1-score} & \multicolumn{1}{c}{$\hat{t}$} & \multicolumn{1}{c}{$\hat{d}$ [ms]} & \multicolumn{1}{c}{FP$_{\hat{d}}$/FP} & \multicolumn{1}{c}{FN$_{\hat{d}}$/FN} \\
\noalign{\smallskip}
\hline
\noalign{\smallskip}
Walking & SVM & \textbf{0.958} & 0.822 & - & 1120 & \textbf{0.214} & \textbf{0.920} \\
 & MLP\_thresh & 0.937 & 0.865 & 0.441 & 10 & 0.693 & 1.211 \\
 & RF\_thresh & 0.928 & 0.854 & 0.379 & 170 & 0.571 & 1.093 \\
 & AutoML\_30 & 0.918 & \textbf{0.882} & - & 10 & 0.673 & 1.082 \\
 \noalign{\smallskip}
\hline
\noalign{\smallskip}
Combined & SVM\_thresh & \textbf{0.847} & \textbf{0.778} & 0.331 & 360 & \textbf{0.486} & \textbf{0.986} \\
 & MLP\_thresh & 0.774 & 0.772 & 0.370 & 10 & 0.734 & 1.087 \\
 & RF\_thresh & 0.720 & 0.738 & 0.327 & 10 & 0.789 & 0.989 \\
 & AutoML\_25200 & 0.707 & 0.752 & - & 10 & 0.768 & 1.025 \\
 \noalign{\smallskip}
\hline
\noalign{\smallskip}
Wheelchair & SVM\_thresh & \textbf{0.622} & \textbf{0.672} & 0.310 & 10 & 0.857 & 1.043 \\
 & MLP\_thresh & 0.414 & 0.540 & 0.376 & 270 & 0.744 & 0.995 \\
 & AutoML\_30 & 0.381 & 0.536 & - & 570 & 1.000 & 0.779 \\
 & RF & 0.170 & 0.289 & - & 970 & \textbf{0.000} & \textbf{0.677} \\
\hline
\end{tabular}
\end{table}

The best model on the self-walking application mode is the SVM with \mbox{95.8 \%} recall. The model's hyperparameters were optimized using a grid search and 5-fold cross validation, resulting in the regularization parameter \(C=0.5\), a polynomial kernel function with degree 2, and \(\gamma = 1 / n_{\mbox{features}}\). The best model for the wheelchair application is also the SVM with \mbox{62.2 \%} recall. The metrics of the best model on the combined dataset reflect an average of the others weighted by the sample size.

\subsection{Limitations}

The recall rate of \mbox{62.2 \%} for the 3-class wheelchair application is much lower than on the self-walking data because there are more complicated and subtle emergency cases. The critical actions of the wheelchair case include: Falling out of the chair and unconsciousness. Especially unconsciousness of a sitting person is very difficult to detect, even for a human. It shows only in a slightly tilted head and sometimes closed eyes. Due to the abstraction of the image feed by OpenPose to single keypoints, it is not possible to distinguish such small changes. However, the classifier is trained on single images of a continuous video stream. Although the metrics show, that not every image is correctly classified, some images of an emergency event are still recognized and result in an emergency call. Nevertheless, an approach with focus on the head position and eye opening might be better suited for this case. In application, this problem is addressed by periodically asking the patient to confirm his or her condition.


\section{Conclusion}

In order to relive hospital staff from time-consuming patient transports, the PeTRA-project develops a mobile robot for autonomous person transport. Currently, most transfers are done in beds to save time, but this does not promote the mobility or autonomy of the patient. To solve this problem, it is possible to be transferred by walking, with a rollator or in a wheelchair. The challenge this paper focuses on, is to ensure the good health of the patient during the autonomous transport. For this, a visual tracking system, Human Pose Estimation with OpenPose, and a classification of the keypoints with recall-optimized models are used to detect emergencies. The problem differs from related work because it considers different modes of transportation for a mobile robot in a highly dynamic environment with multiple people in a scene using only RGB-D data. Difficulties include occlusion by rollators or wheelchairs and additional scenarios such as unconsciousness or crushing hazards. The goal of this paper is to establish a baseline on this new dataset. A subset of the multiclass \textit{PeTRA Emergency Detection Dataset} is published.

We found that in combination with OpenPose keypoints and depth data, the SVM performs best. The methods of threshold moving and time delay helped in optimizing the model to maximize emergency detection and reduce false alarms. For the walking application, we achieve a recall of \mbox{95.8 \%} on single images. In the case of seated transport, the best model is also a SVM with a recall of \mbox{62.2 \%}. Compared to state of the art AutoML approaches, our models perform \mbox{5.3 \%-points} better. A limitation of our approach is the low detection rate of emergencies in the multiclass wheelchair scenario. Due to minor changes in posture or head position, it is very difficult to detect an unconscious patient in a wheelchair.

Further research should focus on end-to-end solutions where OpenPose is substituted. Also, video-based approaches could increase the detection rate compared to single image classification by better accounting for time series behavior. 

\bibliographystyle{spmpsci}
\bibliography{main}

\end{document}